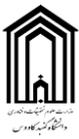
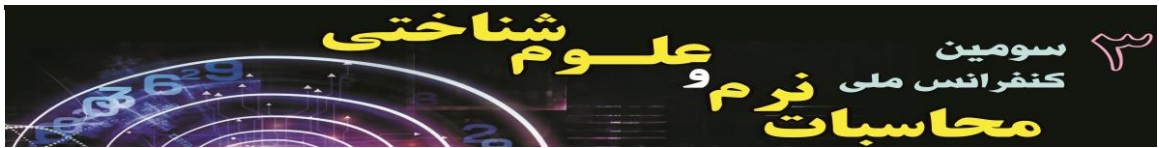

# Improving Object Detection Performance through YOLOv8: A Comprehensive Training and Evaluation Study

**Rana Poureskandar [1] , Shiva Razzagzadeh [2]**

[1] Department of Computer Engineering, Ardabil Branch, Islamic Azad University, Ardabil, Iran;
poureskandar.rana99@gmail.com

[2] Department of Computer Engineering, Ardabil Branch, Islamic Azad University,  Ardabil, Iran;
Shiva.razzaghzadeh@gmail.com

## ABSTRACT

This study evaluated the performance of a YOLOv8-based segmentation model for detecting and segmenting wrinkles in facial images. The model's performance was assessed using standard metrics, including Precision (P), Recall (R), and mean Average Precision (mAP) at thresholds of 0.50 (mAP50) and 0.50–0.95 (mAP50-95), as well as Mask Precision and Mask Recall to evaluate segmentation quality. The model was tested on a validation dataset of 131 images, yielding a Precision of 90.7%, Recall of 89.1%, mAP50 of 87.0%, and mAP50-95 of 10.2%. For segmentation, Mask Precision was 80.7% and Mask Recall was 89.1%. The model performed best in detecting forehead wrinkles, with a Precision of 85.0%, Recall of 80.7%, and mAP50 of 85.7%. Detection of frown lines showed lower performance with a Precision of 80.5% and mAP50 of 81.6%. General wrinkle detection achieved a Precision of 88.6%, but with a lower Recall (81.8%) and mAP50 (83.7%). Although the model demonstrated strong localization and segmentation capabilities, challenges were observed in detecting subtle wrinkles and handling complex lighting or overlapping features, resulting in false positives and under-segmentation in some cases.

## 1. INTRODUCTION

Object detection is a vital component of computer vision. It is essential for enabling interactions between images and text, as well as for monitoring separate things. The capacity of object detection to provide significant insights highlights its numerous applications in various fields, such as machine vision, deep-sea visual monitoring systems, and anomaly detection in medical imaging. The field of deep learning has experienced swift progress in the creation of object detection algorithms [7], [8].

Artificial Intelligence (AI) has created opportunities across various areas, including renewable energy, security, healthcare, and education. The manufacturing industry is particularly positioned for significant automation with Computer Vision (CV). In manufacturing, Quality Inspection (QI) is of paramount importance, since it guarantees clients the integrity and quality of the produced items [9]. Manufacturing presents numerous prospects for automation; yet, obstacles emerge in surface inspection, where flaws may appear in complex patterns. The intricacy of this process renders human-led quality inspection a demanding task, encumbered by challenges including human bias, weariness, expense, and production interruptions [13]. These inefficiencies provide an opportunity for computer vision-based solutions to implement automated quality inspection. These systems can effortlessly integrate into current surface defect detection processes, improving efficiency and avoiding bottlenecks associated with conventional inspection methods [14]. Achieving success necessitates that CV architectures comply with a rigorous set of deployment requirements, which may change across various sectors of the manufacturing industry [15].

In most applications, the goal beyond the basic identification of individual faults; it usually includes the detection of many problems and their precise spatial characteristics [16]. Consequently, the preference favours object detection over image categorisation. The latter exclusively focusses on object identification within a picture, lacking any details regarding their exact location. Object detection architectures can be classified into two main categories: single-stage and two-stage detectors [17]. In two-stage detectors, the detection process is segmented into two phases: feature extraction or proposal, followed by regression and classification to get the final output [18]. This technique provides great accuracy but incurs a substantial computational overhead, making it inefficient for real-time use on resource-limited edge devices. Conversely, single-stage detectors integrate both processes into one phase, enabling classification and regression to transpire simultaneously. This significantly decreases computational demands and offers a more persuasive case for implementation in production settings [19].

YOLOv8 broadens the framework to accommodate various AI tasks, including detection, segmentation, and tracking, hence augmenting its adaptability across many domains. It incorporates a modified CSPDarknet53 backbone and a PAN-FPN neck. YOLOv9 and YOLOv10 incorporate advanced methodologies such as programmable gradient information (PGI) and the generalised efficient layer aggregation network (GELAN), with YOLOv10 obviating the necessity for non-maximum suppression (NMS) via an end-to-end head, thereby enabling real-time object detection [20, 21].

Deep learning models, especially convolutional neural networks (CNNs), have demonstrated significant potential in numerous image identification tasks, including face feature detection. In this context, the YOLO (You Only Look Once) model family, known for its real-time object detection proficiency, has emerged as a favoured option for facial feature detection and segmentation. YOLOv8, the most recent iteration of this architecture, provides enhanced performance regarding speed and accuracy, rendering it an appropriate choice for wrinkle detection and segmentation applications.

This study seeks to assess the efficacy of a YOLOv8-based model specifically designed for the detection and segmentation of face wrinkles. The objective is to utilise advanced deep learning techniques to automate wrinkle identification, enhancing speed, accuracy, and applicability in clinical and commercial environments. The model's proficiency in detecting wrinkles with high precision and recall, along with its segmentation capabilities, will be evaluated through several performance indicators. This assessment will ascertain the feasibility of employing YOLOv8 for automated wrinkle segmentation and its capacity to enable developments in automated face analysis.

This document is organised as follows: The Introduction emphasises the significance of automated wrinkle identification and segmentation across multiple domains, underscoring the capabilities of deep learning models, especially YOLOv8, for this purpose. The Methodology section delineates the architecture and training process of the YOLOv8-based model, as well as the dataset utilised for evaluation. The Results section delineates the model's performance measures, encompassing precision, recall, and mAP, alongside the visual outcomes of the segmentation. The Discussion evaluates the significance of these results, highlighting the model's strengths and shortcomings, whereas the Conclusion encapsulates the findings and proposes avenues for future research.

## 2. Related Work

In computer vision, the precise detection and tracking of cars has become essential for numerous applications, including traffic management and autonomous driving. Technology has progressed from rudimentary detection algorithms to advanced neural networks proficient at executing intricate vehicle recognition jobs across diverse conditions. This literature study examines the evolution of vehicle detection techniques from their inception to the contemporary breakthroughs in YOLO, emphasising key contributions and innovations that have influenced modern vehicle monitoring systems.

Vehicle detection and tracking are essential elements of computer vision monitoring systems, enabling functions such as vehicle enumeration, accident detection, traffic pattern analysis, and surveillance. Vehicle detection encompasses the identification and localisation of vehicles via bounding boxes, whereas tracking comprises monitoring and forecasting vehicle movements through trajectories [22]. Initially, convolutional algorithms concentrated on background elimination and user-defined feature extraction; however, they encountered challenges with dynamic backdrops and fluctuating weather conditions [23]. Barth et al. established the Stixels approach in their research to overcome these challenges, employing colour schema to translate movement information [24]. Convolutional neural networks (CNNs) have been utilised to address challenges like as occluding objects, diverse backdrops, and latency issues, hence improving accuracy. Numerous studies have investigated CNN architectures designed for these tasks, such as RCNN, FRCNN, SSD, and ResNet [7,25,26,27,28,29,30,31,32,33]. Furthermore, Azimjonov and Özmen conducted a comparative analysis of classical machine learning and deep learning algorithms for the detection of road vehicles [11]. Vehicle tracking approaches encompass detection tracking via bounding boxes and appearance-based tracking emphasising visual characteristics. The integration of YOLO for detection and a CNN for tracking has exhibited superior performance relative to nine other machine learning models, indicating a viable methodology for vehicle monitoring systems.

YOLO has markedly enhanced the precision of vehicle identification by treating it as a regression task through the utilisation of convolutional neural networks (CNNs), effectively identifying the locations, types, and confidence scores of vehicles. It improves detection velocity while mitigating motion blur by supplying bounding boxes and class probabilities. Figure 1 presents a detailed visual representation of the incremental improvements in the YOLO architecture series from YOLOv1 to YOLOv10. YOLOv2, utilising GPU capabilities and the anchor box methodology, enhanced its predecessor in the detection, classification, and tracking of vehicles [34]. Ćorović, Ilić et al. introduced YOLOv3, which was trained on five categories: automobiles, trucks, street signs, individuals, and traffic signals. It was proposed to detect traffic participants effectively across various weather conditions [35]. YOLOv4 aimed to improve the detection speeds of slow-moving vehicles in video feeds [36], whereas YOLOv5 employed an infrared camera to identify heavy vehicles in snowy conditions, enabling real-time parking space prediction due to its efficient architecture and rapid identification capabilities [37].

## 3. Material and methods

This section outlines the methodologies and tools used for the detection and segmentation of wrinkles using YOLOv8. It includes details about the dataset, preprocessing steps, model architecture, training methodology, evaluation metrics, and limitations with potential future improvements.

### 3.1. Dataset

The dataset used in this study was sourced from the **Roboflow** platform, specifically tailored for facial wrinkle detection and segmentation. It comprises high-resolution images of human faces annotated with labels for specific regions such as the forehead, frown lines, and general wrinkles. The dataset was split into three subsets: 70% for training, 20% for validation during training, and 10% for testing the model's performance after training. The dataset features diverse samples with variations in age, skin tone, and wrinkle intensity to ensure the model's robustness across different demographics.

### 3.2. Data Preprocessing

Before feeding the images into the model, several preprocessing steps were performed to prepare the data. All images were resized to a standard resolution of 640×640 pixels to ensure uniformity and reduce computational complexity. Pixel values were normalized to a range of 0 to 1, which is crucial for faster convergence during training. Additionally, data augmentation techniques such as random flipping, cropping, rotation, and brightness adjustments were applied to artificially increase the dataset's diversity. This helps the model generalize better and minimizes the risk of overfitting.

### 3.3. Model Architecture

The YOLOv8 segmentation model was chosen for its efficiency and accuracy in segmentation tasks. Specifically, the **YOLOv8s-seg** variant was employed due to its lightweight architecture, which balances computational efficiency and performance. The model comprises three primary components: a backbone for extracting hierarchical features from input images, a neck to combine and refine multi-scale features, and a head that generates bounding boxes, segmentation masks, and class probabilities. Unlike traditional detection models, this segmentation-specific architecture allows precise mask predictions for the wrinkle regions, enabling accurate localization and segmentation simultaneously.

### 3.4. Model Training and Evaluation

The training process utilized the YOLOv8 framework with pretrained weights derived from the COCO dataset to leverage transfer learning. The model was trained for 50 epochs using an adaptive learning rate scheduler and the Adam optimizer, which ensures stable convergence. A batch size appropriate for GPU memory capacity was selected. The model's performance was evaluated using metrics such as Precision (P), Recall (R), and mean Average Precision (mAP) for both bounding boxes and segmentation masks. These metrics provide a comprehensive understanding of the model's capability to accurately detect and segment wrinkles across various images.

The confusion matrix is a common tool used to analyze the relationship between actual and predicted values, encompassing true positives (TP), false positives (FP), false negatives (FN), and true negatives (TN) [38-48]. The layout of the confusion matrix is shown in **Table 1**. To assess the classification performance of all machine learning algorithms, we typically employ various metrics such as mean squared error (MSE), F1-score, accuracy, precision, and recall (sensitivity). These metrics offer a thorough evaluation of the models' effectiveness in accurately classifying diabetes mellitus cases.

**Table 1.    Structure of Confusion Matrix.**

|  |  | Actual Class |  |
|---|---|---|---|
|  |  | Negative | Positives |
| Predicted | Negative | **TP (True Positive)** | FP (False positive) |
| Class | Positives | FN (False Negative) | **TN (True Negative)** |

True positives (TP) and true negatives (TN) indicate the counts of correctly identified positive and negative samples, respectively. As outlined in [38-48], accuracy is defined as the ratio of the number of correct predictions (both TP and TN) to the total number of predictions made, which includes all positive cases, accounting for true positives and any falsely identified positives. Recall, or sensitivity, represents the ratio of correctly identified positive cases to all actual positive cases, combining true positives with false negatives (FN) to evaluate the model's ability to identify all positive instances. In contrast, precision is defined as the ratio of correctly identified positive cases to all predicted positive cases, encompassing both true positives and false positives (FP) [38-48].

$$Accuracy = \frac{TP + TN}{TP + FP + TN + FN} \tag{1}$$

$$Precision = \frac{TP}{TP + FP} \tag{2}$$

$$\text{Re}\,call = \frac{TP}{TP + FN} \tag{3}$$

$$F1 - score = 2 * \frac{\Pr ecision * \text{Re}\,call}{\Pr ecision * \text{Re}\,call} \tag{4}$$

Where TP = true positive, TN = true negative, FP = false positive, and FN = false negative.

### 3.5. Evaluation and Results

The evaluation of the YOLOv8-based segmentation model was conducted using standard performance metrics to assess its ability to detect and segment wrinkles accurately. These metrics included **Precision (P)**, **Recall (R)**, and **mean Average Precision (mAP)** at thresholds of 0.50 (mAP50) and 0.50–0.95 (mAP50-95). Additionally, specific segmentation-related metrics such as **Mask Precision** and **Mask Recall** were analyzed to evaluate the quality of the predicted segmentation masks.

The model was validated on a separate validation dataset consisting of 131 images. The overall detection results showed a **Precision of 90.7%** and a **Recall of 89.1%**, with an mAP50 of 87.0% and an mAP50-95 of 10.2%. Segmentation-specific evaluation revealed a Mask Precision of 80.7% and a Mask Recall of 89.1%, indicating a moderate capability for accurate segmentation. When broken down by classes, the model demonstrated its highest performance in detecting **forehead wrinkles**, achieving a Precision of 85.0%, a Recall of 80.7%, and an mAP50 of 85.7%. For **frown lines**, the model showed lower performance with a Precision of 80.5% and an mAP50 of 81.6%. Finally, the detection of **general wrinkles** had a Precision of 88.6% but lower Recall (81.8%) and mAP50 (83.7%). The model's predictions were visualized on test images, showcasing its ability to localize and segment wrinkle regions effectively. However, it struggled with subtle wrinkles and regions with complex lighting or overlapping features, leading to false positives and under-segmentation in certain cases.

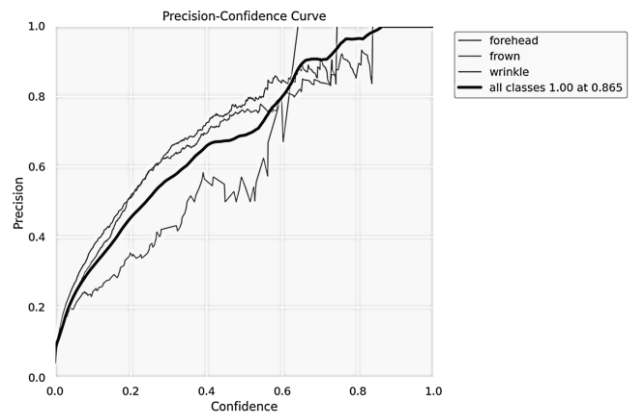

**Figure 1. Precision-Confidence curve visualizes the relationship between the model's confidence in its predictions and its precision**

The Precision-Confidence curve visualizes the relationship between the model's confidence in its predictions and its precision. Each line represents a different class: forehead, frown, and wrinkle. The x-axis indicates the model's confidence level, ranging from 0 to 1, with 1 representing the highest confidence. The y-axis shows the corresponding precision. A higher precision score indicates that a higher proportion of positive predictions made by the model are actually correct. The blue line, representing the average precision across all classes, shows that the model achieves its highest precision of 1.00 at a confidence level of 0.865. This suggests that when the model is highly confident in its predictions, it is almost always correct.

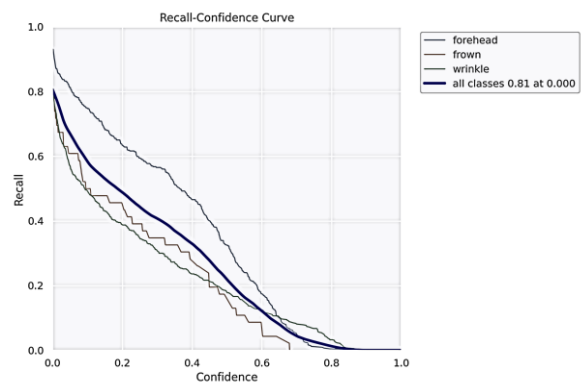

**Figure 2. Recall-Confidence curve visualizes the relationship between the model's confidence in its predictions and its recall**

The Recall-Confidence curve visualizes the relationship between the model's confidence in its predictions and its recall. Each line represents a different class: forehead, frown, and wrinkle. The x-axis indicates the model's confidence level, ranging from 0 to 1, with 1 representing the highest confidence. The y-axis shows the corresponding recall. A higher recall score indicates that a higher proportion of actual positive cases were identified correctly by the model. The blue line, representing the average recall across all classes, shows that the model achieves its highest recall of 0.81 at a confidence level of 0.000. This suggests that when the model is very confident in its predictions, it is able to identify a high proportion of the actual positive cases.

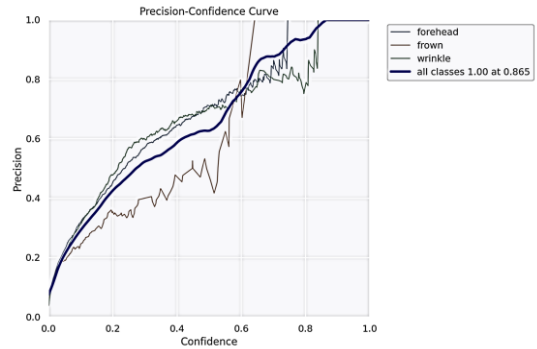

**Figure 3. Precision-Confidence curve visualizes the relationship between the model's confidence in its predictions and its precision.**

The Precision-Confidence curve visualizes the relationship between the model's confidence in its predictions and its precision. Each line represents a different class: forehead, frown, and wrinkle. The x-axis indicates the model's confidence level, ranging from 0 to 1, with 1 representing the highest confidence. The y-axis shows the corresponding precision. A higher precision score indicates that a higher proportion of positive predictions made by the model are actually correct. The blue line, representing the average precision across all classes, shows that the model achieves its highest precision of 1.00 at a confidence level of 0.865. This suggests that when the model is highly confident in its predictions, it is almost always correct.

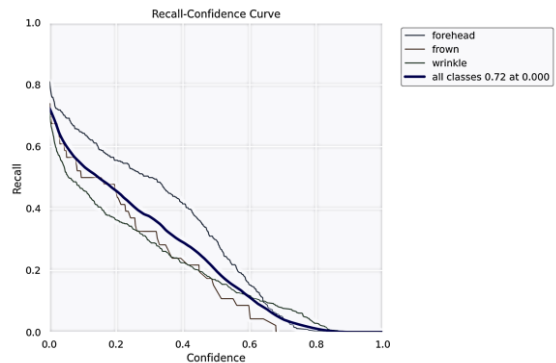

**Figure 4. Recall-Confidence curve visualizes the relationship between the model's confidence in its predictions and its recall.**

The Recall-Confidence curve visualizes the relationship between the model's confidence in its predictions and its recall. Each line represents a different class: forehead, frown, and wrinkle. The x-axis indicates the model's confidence level, ranging from 0 to 1, with 1 representing the highest confidence. The y-axis shows the corresponding recall. A higher recall score indicates that a higher proportion of actual positive cases were identified correctly by the model. The blue line, representing the average recall across all classes, shows that the model achieves its highest recall of 0.72 at a confidence level of 0.000. This suggests that when the model is very confident in its predictions, it is able to identify a high proportion of the actual positive cases.

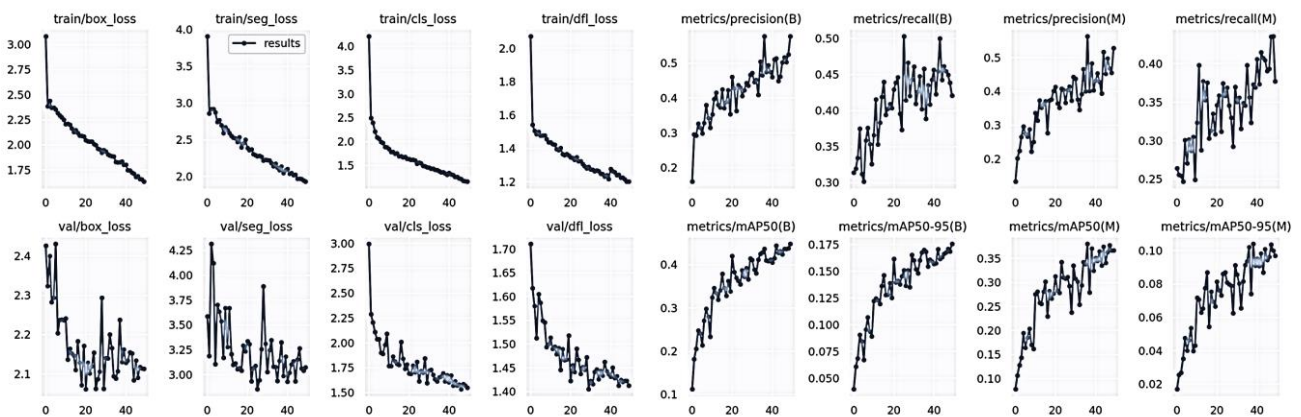

**Figure 5. provided graph illustrates the training and validation loss of a deep learning model over 50 epochs.**

The provided graph illustrates the training and validation loss of a deep learning model over 50 epochs. The model's performance is evaluated using various metrics such as box loss, segmentation loss, classification loss, and detection loss. The metrics for precision and recall are also plotted for both the bounding box and mask predictions. The decreasing trend in both training and validation losses indicates that the model is learning effectively. However, there seems to be a slight overfitting towards the end of training as the validation loss starts to increase. Overall, the model appears to be performing well, with improving performance on most metrics.

During the training process of the model, a systematic reduction in loss values and improvement in evaluation metrics are observed, reflecting the model's gradual learning and optimization. Initially, the **Box Loss**, which measures the error in predicted bounding box coordinates, is relatively high. This is expected, as the model starts with random predictions and lacks spatial awareness. Similarly, **Class Loss**, representing errors in object classification, and **Objectness Loss**, indicating inaccuracies in predicting object presence, also begin at high levels. Over time, as the model is exposed to training data and updates its parameters, these losses decrease steadily, demonstrating the model's improving ability to detect, localize, and classify objects.

Concurrently, evaluation metrics like **mean Average Precision (mAP)** provide insight into the model's performance from a practical perspective. Metrics such as **mAP**, which measures accuracy at an IoU threshold of 0.5, and **mAP**, which evaluates across a range of IoU thresholds, show continuous improvement throughout training. This trend indicates that the model is not only learning to make accurate predictions but also achieving a good balance between precision (minimizing false positives) and recall (minimizing false negatives).

These changes suggest that the training process is effectively guiding the model toward convergence. The optimization algorithms adjust the model's weights to minimize the total loss, while regular evaluation ensures that the model generalizes well to unseen data. Improvements in the mAP metrics, in particular, highlight the model's ability to detect objects with both high confidence and accuracy across various sizes and contexts in the dataset.

In summary, the training process results in a progressively better-performing model. Loss reduction reflects improved learning at a granular level, while mAP improvement provides a clear picture of practical performance in object detection tasks. Together, these metrics confirm the success of the training strategy and the model's ability to generalize effectively for real-world applications.

### 3.6. Model Limitations and Future Work

While the YOLOv8-based segmentation model achieved promising results, it also exhibited several limitations. The dataset size, although diverse, was relatively small, which might restrict the model's ability to generalize to unseen scenarios, such as uncommon lighting conditions or overlapping wrinkles. The model's mAP values at higher IoU thresholds were suboptimal, suggesting a need for improved localization precision. In the future, expanding the dataset to include a wider range of demographics, age groups, and environmental conditions could address these issues. Additionally, exploring advanced architectures, such as Transformer-based models, may enhance segmentation accuracy. Optimizing the model for real-time applications and integrating additional modalities, like 3D imaging, could further expand its potential use cases, such as in dermatological diagnostics or personalized skincare recommendations.

## 4. Discussion

The results observed during the training and evaluation phases provide valuable insights into the strengths and limitations of the model. The steady reduction in **Box Loss**, **Class Loss**, and **Objectness Loss** demonstrates that the model effectively learns spatial and categorical relationships within the dataset. However, the convergence patterns of these losses can also reveal potential challenges, such as overfitting if the validation loss stagnates or increases while the training loss continues to decrease. Regular monitoring of these trends is critical to ensuring the model generalizes well to unseen data.

Evaluation metrics like **mAP** further emphasize the model's practical performance. While consistent improvements in these metrics are promising, their growth rate may plateau, indicating diminishing returns from additional training epochs. This plateau could result from the dataset's complexity or limitations in the model's architecture. It's essential to consider strategies such as data augmentation, hyperparameter tuning, or architecture refinement to address such challenges.

Another key discussion point is the balance between **precision** and **recall**, which affects the model's reliability in different applications. High precision with lower recall might be suitable for scenarios requiring fewer false positives, while high recall might be preferable for applications where missing detections is unacceptable. Fine-tuning the decision threshold for classification can help optimize this trade-off according to specific use cases. Lastly, real-world deployment often introduces challenges such as domain shifts or unseen object variations. While the training metrics and evaluations provide a baseline understanding, further testing on diverse datasets and under varied conditions is crucial to ensure robustness. These discussions underscore the need for iterative improvements and adaptability in the training and evaluation pipeline.

## 5. Conclusion

In conclusion, the training and evaluation process has demonstrated the model's ability to detect, localize, and classify objects with increasing accuracy. The consistent reduction in loss values highlights the effectiveness of the optimization process, while improvements in mAP metrics affirm the model's readiness for practical applications. These results validate the design choices in the model architecture and the training strategy. However, the observed challenges, such as potential overfitting and performance plateaus, emphasize the need for further refinements. Enhancements in data diversity, more advanced augmentation techniques, and hyperparameter optimization could help push the model toward better generalization and adaptability. Additionally, integrating feedback from real-world deployments can guide future iterations of the model for more robust and reliable performance. The findings from this work contribute to a better understanding of object detection model training and provide a solid foundation for continued exploration. By addressing the discussed limitations and leveraging advancements in machine learning, the model can be adapted to meet specific requirements across various domains, making it a versatile and impactful tool.

## REFERENCES

[1] M. Safaldin, N. Zaghden and M. Mejdoub, "Moving object detection based on enhanced YOLO-V2 model", Proc. 5th Int. Congr. Human-Computer Interact. Optim. Robotic Appl. (HORA), pp. 1-8, Jun. 2023.

[2] S. Ammar, T. Bouwmans, N. Zaghden and M. Neji, "Deep detector classifier (DeepDC) for moving objects segmentation and classification in video surveillance", IET Image Process., vol. 14, no. 8, pp. 1490-1501, Jun. 2020.

[3] S. Ammar, T. Bouwmans, N. Zaghden and N. Mahmoud, "From moving objects detection to classification and recognition: A review for smart environments", Proc. Towards Smart World, pp. 289-316, 2020.

[4] E. M. Ibrahim, M. Mejdoub and N. Zaghden, "Semantic analysis of moving objects in video sequences", Proc. Int. Conf. Emerg. Technol. Intell. Syst., pp. 257-269, 2022.

[5] F. Ben Aissa, M. Hamdi, M. Zaied and M. Mejdoub, "An overview of GAN-DeepFakes detection: Proposal improvement and evaluation", Multimedia Tools Appl., vol. 83, no. 11, pp. 32343-32365, Sep. 2023.

[6] H. Ma, T. Celik and H. Li, "Fer-YOLO: Detection and classification based on facial expressions", Proc. Image Graphics: 11th Int. Conf., pp. 28-39, 2021.

[7] D. Feng, A. Harakeh, S. L. Waslander and K. Dietmayer, "A review and comparative study on probabilistic object detection in autonomous driving", IEEE Trans. Intell. Transp. Syst., vol. 23, no. 8, pp. 9961-9980, Aug. 2022.

[8] K. Tong, Y. Wu and F. Zhou, "Recent advances in small object detection based on deep learning: A review", Image Vis. Comput., vol. 97, May 2020.

[9] M. Hussain, H. Al-Aqrabi, M. Munawar, R. Hill and S. Parkinson, "Exudate regeneration for automated exudate detection in retinal fundus images", IEEE Access, vol. 11, pp. 83934-83945, 2022.

[10] M. Hussain, M. Dhimish, V. Holmes and P. Mather, "Deployment of AI-based RBF network for photovoltaics fault detection procedure", AIMS Electron. Electr. Eng., vol. 4, no. 1, pp. 1-18, 2020.

[11] S. A. Singh and K. A. Desai, "Automated surface defect detection framework using machine vision and convolutional neural networks", J. Intell. Manuf., vol. 34, no. 4, pp. 1995-2011, Apr. 2023.

[12] D. Weichert, P. Link, A. Stoll, S. Rüping, S. Ihlenfeldt and S. Wrobel, "A review of machine learning for the optimization of production processes", Int. J. Adv. Manuf. Technol., vol. 104, no. 5, pp. 1889-1902, Oct. 2019.

[13] J. Wang, Y. Ma, L. Zhang, R. X. Gao and D. Wu, "Deep learning for smart manufacturing: Methods and applications", J. Manuf. Syst., vol. 48, pp. 144-156, Jul. 2018.

[14] D. Weimer, B. Scholz-Reiter and M. Shpitalni, "Design of deep convolutional neural network architectures for automated feature extraction in industrial inspection", CIRP Ann., vol. 65, no. 1, pp. 417-420, 2016.

[15] S. Kulik and A. Shtanko, "Experiments with neural net object detection system YOLO on small training datasets for intelligent robotics", Proc. Adv. Technol. Robot. Intell. Syst. ITR, pp. 157-162, 2020.

[16] J. Yang, S. Li, Z. Wang, H. Dong, J. Wang and S. Tang, "Using deep learning to detect defects in manufacturing: A comprehensive survey and current challenges", Materials, vol. 13, no. 24, pp. 5755, Dec. 2020.

[17] P. Soviany and R. T. Ionescu, "Optimizing the trade-off between single-stage and two-stage deep object detectors using image difficulty prediction", Proc. 20th Int. Symp. Symbolic Numeric Algorithms for Scientific Comput. (SYNASC), pp. 209-214, Sep. 2018.

[18] L. Du, R. Zhang and X. Wang, "Overview of two-stage object detection algorithms", J. Phys. Conf. Ser., vol. 1544, no. 1, May 2020.

[19] F. Sultana, A. Sufian and P. Dutta, "A review of object detection models based on convolutional neural network" in Intelligent Computing: Image Processing Based Applications, Kolkata, India:Springer, pp. 1-16, 2020.

[20] Jocher, G.; Munawar, M.R.; Chaurasia, A. YOLO: A Brief History; 2023. Available online: https://www.scirp.org/reference/referencespapers?referenceid=3532980 (accessed on 11 February 2023).

[21] Al Rabbani Alif, M.; Hussain, M. YOLOv1 to YOLOv10: A comprehensive review of YOLO variants and their application in the agricultural domain. arXiv 2024, arXiv:2406.10139.

[22] Fernandez-Sanjurjo, M.; Bosquet, B.; Mucientes, M.; Brea, V.M. Real-time visual detection and tracking system for traffic monitoring. Eng. Appl. Artif. Intell. 2019, 85, 410–420.

[23] Mandellos, N.A.; Keramitsoglou, I.; Kiranoudis, C.T. A background subtraction algorithm for detecting and tracking vehicles. Expert Syst. Appl. 2011, 38, 1619–1631.

[24] Erbs, F.; Barth, A.; Franke, U. Moving vehicle detection by optimal segmentation of the dynamic stixel world. In Proceedings of the 2011 IEEE Intelligent Vehicles Symposium (IV), Baden-Baden, Germany, 5–9 June 2011; pp. 951–956.

[25] Ren, S.; He, K.; Girshick, R.; Sun, J. Faster R-CNN: Towards real-time object detection with region proposal networks. IEEE Trans. Pattern Anal. Mach. Intell. 2016, 39, 1137–1149.

[26] Liu, W.; Anguelov, D.; Erhan, D.; Szegedy, C.; Reed, S.; Fu, C.Y.; Berg, A.C. Ssd: Single shot multibox detector. In Proceedings of the Computer Vision–ECCV 2016: 14th European Conference, Amsterdam, The Netherlands, 11–14 October 2016; Proceedings, Part I 14. Springer: Berlin/Heidelberg, Germany, 2016; pp. 21–37.

[27] He, K.; Zhang, X.; Ren, S.; Sun, J. Deep residual learning for image recognition. In Proceedings of the IEEE Conference on Computer Vision and Pattern Recognition, Las Vegas, NV, USA, 26 June–1 July 2016; pp. 770–778.

[28] Alif, M.A.R.; Hussain, M.; Tucker, G.; Iwnicki, S. BoltVision: A Comparative Analysis of CNN, CCT, and ViT in Achieving High Accuracy for Missing Bolt Classification in Train Components. Machines 2024, 12, 93.

[29] Alif, M.A.R.; Hussain, M. Lightweight Convolutional Network with Integrated Attention Mechanism for Missing Bolt Detection in Railways. Metrology 2024, 4, 254–278.

[30] Alif, M.A.R. Attention-Based Automated Pallet Racking Damage Detection. Int. J. Innov. Sci. Res. Technol. 2024, 9, 728–740.

[31] Hussain, M. YOLO-v5 Variant Selection Algorithm Coupled with Representative Augmentations for Modelling Production-Based Variance in Automated Lightweight Pallet Racking Inspection. Big Data Cogn. Comput. 2023, 7, 120.

[32] Zahid, A.; Hussain, M.; Hill, R.; Al-Aqrabi, H. Lightweight convolutional network for automated photovoltaic defect detection. In Proceedings of the 2023 9th International Conference on Information Technology Trends (ITT), Dubai, United Arab Emirates, 24–25 May 2023; IEEE: New York, NY, USA, 2023; pp. 133–138.

[33] Alif, M.A.R. State-of-the-Art Bangla Handwritten Character Recognition Using a Modified Resnet-34 Architecture. Int. J. Innov. Sci. Res. Technol. 2024, 9, 438–448.

[34] Sang, J.; Wu, Z.; Guo, P.; Hu, H.; Xiang, H.; Zhang, Q.; Cai, B. An improved YOLOv2 for vehicle detection. Sensors 2018, 18, 4272. [Google Scholar] [CrossRef]

[35] Ćorović, A.; Ilić, V.; Đurić, S.; Marijan, M.; Pavković, B. The real-time detection of traffic participants using YOLO algorithm. In Proceedings of the 2018 26th Telecommunications Forum (TELFOR), Belgrade, Serbia, 20–21 November 2018; pp. 1–4.

[36] Hu, X.; Wei, Z.; Zhou, W. A video streaming vehicle detection algorithm based on YOLOv4. In Proceedings of the 2021 IEEE 5th Advanced Information Technology, Electronic and Automation Control Conference (IAEAC), Chongqing, China, 12–14 March 2021; Volume 5, pp. 2081–2086.

[37] Kasper-Eulaers, M.; Hahn, N.; Berger, S.; Sebulonsen, T.; Myrland, Ø.; Kummervold, P.E. Detecting heavy goods vehicles in rest areas in winter conditions using YOLOv5. Algorithms 2021, 14, 114.

[38] Abdollahi, J., & Aref, S. (2024). Early Prediction of Diabetes Using Feature Selection and Machine Learning Algorithms. SN Computer Science, 5(2), 217.

[39] Javadzadeh Barzaki, M. A., Negaresh, M., Abdollahi, J., Mohammadi, M., Ghobadi, H., Mohammadzadeh, B., & Amani, F. (2022, July). USING DEEP LEARNING NETWORKS FOR CLASSIFICATION OF LUNG CANCER NODULES IN CT IMAGES. In Iranian Congress of Radiology (Vol. 37, No. 2, pp. 34-34). Iranian Society of Radiology.

[40] Abdollahi, J., & Mehrpour, O. (2024, February). Using Machine Learning Algorithms for Coronary Artery Disease (CAD) Prediction Prediction of Coronary Artery Disease (CAD) Using Machine Learning Algorithms. In 2024 10th International Conference on Artificial Intelligence and Robotics (QICAR) (pp. 164-172). IEEE.

[41] Barzaki, M. A. J. Z., Abdollahi, J., Negaresh, M., Salimi, M., Zolfaghari, H., Mohammadi, M., ... & Amani, F. (2023, November). Using Deep Learning for Classification of Lung Cancer on CT Images in Ardabil Province: Classification of Lung Cancer using Xception. In 2023 13th International Conference on Computer and Knowledge Engineering (ICCKE) (pp. 375-382). IEEE.

[42] Abdollahi, J., & Amani, F. (2024, February). The impact of analysis Suicide using machine learning algorithms in Ardabil: A performance analysis: using machine learning algorithms in analysis Suicide. In 2024 10th International Conference on Artificial Intelligence and Robotics (QICAR) (pp. 24-29). IEEE.

[43] Amani, F., Abdollahi, J., & Amani, P. (2024, February). Identify the Factors Influencing Suicide among Ardabil city People Using Feature Selection: Identify the Factors Influencing Suicide among Ardabil using machine learning. In 2024 10th International Conference on Artificial Intelligence and Robotics (QICAR) (pp. 17-23). IEEE.

[44] Nouri-Moghaddam, B., Shahabian, M. I., & Naji, H. R. Multi-Agent Based PGP Architecture. architecture, 3, 5.

[45] Nouri-Moghaddam, B., & Naji, H. R. (2013, May). Improving HBQ Authentication and Access control in wireless sensor network. In The 5th Conference on Information and Knowledge Technology (pp. 82-87). IEEE.

[46] Tarif, M., Mirzaei, A., & Nouri-Moghaddam, B. (2024). Optimizing RPL Routing Using Tabu Search to Improve Link Stability and Energy Consumption in IoT Networks. arXiv preprint arXiv:2408.06702.

[47] Nouri-Moghaddam, B., & Naji, H. R. (2015). A novel authentication and access control framework in wireless sensor networks. Journal of Advanced Computer Science and Technology, 4(1), 122-135.

[48] Nouri-Moghaddam, B., Ghazanfari, M., & Fathian, M. (2020). A novel filter-wrapper hybrid gene selection approach for microarray data based on multi-objective forest optimization algorithm. Decision Science Letters, 9(3), 271-290.